\documentclass[letterpaper, 10 pt, conference]{ieeeconf}  

\IEEEoverridecommandlockouts                              
\usepackage[utf8]{inputenc}
\usepackage[T1]{fontenc}
\usepackage{cite}
\usepackage{amsmath,amssymb,amsfonts}
\usepackage{algorithmic}
\usepackage{graphicx}
\usepackage{textcomp}
\usepackage{xcolor}
\usepackage{hyperref}
\usepackage{subcaption}
\usepackage{adjustbox}

\title{\LARGE \bf
Contactless hand tremor amplitude measurement using smartphones: development and pilot evaluation
}

\author{James Bungay$^{1}$, Osasenaga Emokpae$^{1}$, Samuel D. Relton$^{2}$, Jane Alty$^{3}$, Stefan Williams$^{4}$, Hui Fang$^{5}$\\
and David C Wong$^{1}$
\thanks{This work is sponsored by the National Institute for Health Research (NIHR203399)}
\thanks{$^{1}$J Bungay, O Emokpae, and DC Wong are with the Department of Computer Science,
        University of Manchester, UK
        {\tt\small corresponding author: d.c.wong@leeds.ac.uk}}%
\thanks{$^{2}$SD Relton is with the Leeds Institute of Health Sciences, University of Leeds, UK,
        {\tt\small s.d.relton@leeds.ac.uk}}%
\thanks{$^{3}$J Alty is with the University of Tasmania, Australia, University of Leeds, UK,
        {\tt\small jane.alty@utas.edu.au}}%
\thanks{$^{3}$S Williams is with Leeds Teaching Hospitals Trust, UK,
        {\tt\small stefan.williams2@nhs.net}}%
\thanks{$^{3}$H Fang is with the Department of Computer Science, Loughborough University, UK,
        {\tt\small h.fang@lboro.ac.uk}}%
}

\begin{document}

\maketitle
\thispagestyle{empty}
\pagestyle{empty}

\begin{abstract}

Background -  Physiological tremor is defined as an involuntary and rhythmic shaking. Tremor of the hand is a key symptom of multiple neurological diseases, and its frequency and amplitude differs according to both
disease type and disease progression. In routine clinical practice, tremor frequency and amplitude are assessed by expert rating using a 0 to 4 integer scale. Such ratings are subjective and have poor inter-rater reliability. There is thus a clinical need for a practical and accurate method for objectively assessing hand tremor. 

Objective - to develop a proof-of-principle method to measure hand tremor amplitude from smartphone videos.

Methods - We created a computer vision pipeline that automatically extracts salient points on the hand and produces a 1-D time series of movement due to tremor, in pixels. Using the smartphones' depth measurement, we convert this measure into real distance units. We assessed the accuracy of the method using 60 videos of simulated tremor of different amplitudes from two healthy adults. Videos were taken at distances of 50, 75 and 100 cm between hand and camera. The participants had skin tone II and VI on the Fitzpatrick scale. We compared our method to a gold-standard measurement from a slide rule. Bland-Altman methods agreement analysis indicated a bias of 0.04 cm and 95\% limits of agreement from -1.27 to 1.20 cm. Furthermore, we qualitatively observed that the method was robust to differences in skin tone and limited occlusion, such as a band-aid affixed to the participant's hand.

Clinical relevance - We have demonstrated how tremor amplitude can be measured from smartphone videos. In conjunction with tremor frequency, this approach could be used to help diagnose and monitor neurological diseases.

\end{abstract}

\section{Introduction}
Hand tremor is a common symptom of multiple diseases, including Parkinson's disease, essential tremor, and multiple sclerosis. Assessment of tremor activity is an important clinical task that can help in diagnosis of disease and evaluating response to treatment.

Tremor is assessed clinically by considering its frequency and amplitude. The standard clinical methods of measuring both tremor amplitude and frequency are subjective. A clinician visually observes a patient tremor and makes an estimate of both measures, categorising it with a severity rating \cite{goetz2008movement, elble2016essential}. Such visual estimates of movement disorders are usually performed in face-to-face consultations, and there is large inter-rater variability between expert clinicians such that tremor diagnoses are frequently incorrect \cite{jain2006common, bajaj2010accuracy}

Objective measurements of tremor frequency is possible using an accelerometer strapped to the hand \cite{vial2019electrophysiological}. Tremor amplitude is rarely derived directly, and instead, the amplitude of the acceleration signal is taken as a proxy for the amplitude of the displacement. The practical use of accelerometers is limited in two respects; it requires non-standard specialist equipment, and also adds weight to the hand in a way that may alter tremor characteristics.

Instead, it may be possible to derive tremor frequency and amplitude measurements directly from smartphone videos. Analysis of smartphones videos has been used to assess other biomarkers of neurological conditions \cite{williams2020supervised, zhao2020time}. The ubiquity of smartphones means that such computer vision approaches have the potential to be used in multiple contexts. For example, they could be used for remote consultations or for monitoring of disease progression. Recently, we have developed a method for extracting tremor \textit{frequency} directly from smartphone video recordings of hand tremor \cite{williams2021accuracy}.

Here, we propose and demonstrate proof-of-principle for a method that enables measurement of tremor \textit{amplitude} from smartphone videos.

\section{Method}
\subsection{Technical Description}
\begin{figure*}[htbp]
    \centerline{\includegraphics[width=0.9\textwidth]{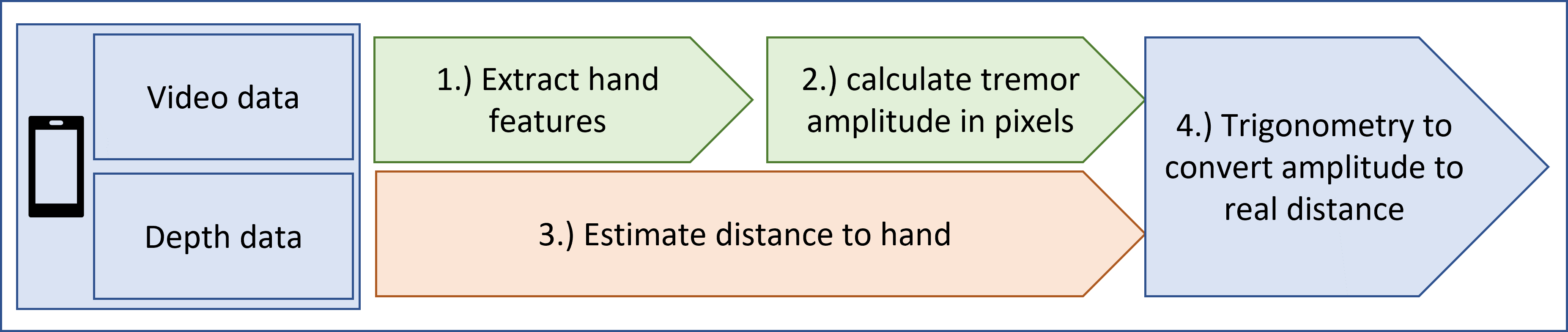}}
    \caption{Overview of hand-tremor amplitude measurement using smartphone video analysis}
    \label{fig:schematic}
\end{figure*}
The method consists of four main parts, shown in figure \ref{fig:schematic}. We assume that data has been collected via a modern smartphone that can capture both video and a depth measurement at the centre of the camera frame. In our work, we used an iPhone XR to provide both measurements.

\textbf{1. Extract hand features} - Using the video data, we extracted salient points on the hand using the Mediapipe hand tracker \cite{zhang2020mediapipe}. This two-stage process identifies the palm region using a U-net and then fits salient points that are consistent with pre-specified hand pose model. The hand tracker returns a tuple of $\{x,y,t\}$ corresponding to the $x$ and $y$ pixels, and time. For robustness, we monitored movement over multiple points, corresponding to the base (metacarpal), middle (interphalangeal) and tip of thumb and forefinger.

\textbf{2. Calculate tremor amplitude in pixels} - In our controlled scenarios, the primary direction of tremor was horizontal. We therefore used only $\{x,t\}$ to represent the motion waveform over time. To filter out low frequency motion due to gross arm movements, we processed the waveform by first extracting peaks and troughs. We used a simple forward-difference to estimate the gradient and located zero crossings. The difference in $x$ between adjacent zero crossings was an instantaneous estimate of the amplitude - this was calculated for all adjacent pairs of zero crossings. From this set, we used the median value to be robust against artificial increase in tremor due to `ramp-up' of the tremor motion from an at-rest state.

\begin{table}[h]
\centering
\begin{tabular}{|c|c|c|}
 \hline
 Camera Property & Symbol & Value \\
 \hline
 \hline
 Physical Lens Focal Length & $f$ & 2.87 mm \\
 \hline
 35mm Equiv. Lens Focal Length & $f_e$ & 32 mm \\
 \hline
 Sensor Aspect Ratio & $a_s$ & 0.75 (i.e. 3:4) \\
 \hline
 Video Aspect Ratio & $a_v$ & 0.5625 (i.e. 9:16) \\
 \hline
 Horizontal Video Resolution & $r_h$ & 1080 px \\ 
 \hline
\end{tabular}
\caption{Specifications of the iPhone XR front facing camera.}
\label{table:camera-specs}
\end{table}

\textbf{3. Convert amplitude from pixels to distance units:} Finally, we converted the pixel distance into true distance units. The conversion relies on the distance between the smartphone and the hand. We measured this distance using the Apple TrueDepth sensor, using the front-facing camera on an iPhone XR. In controlled experiments, we assessed the accuracy of the depth sensor by reading the TrueDepth sensor distance to an object at known depths. In these experiments, the camera was affixed to a tripod in good lighting conditions. At a known distance of 40cm, the root mean squared error over six sensor measurements was 0.12 cm. At 100 cm, this error increased slightly to 0.38 cm. 

The conversion also requires knowledge of physical characteristics of the camera described in Table \ref{table:camera-specs}. For a video shot in portrait, the horizontal sensor size of the camera, $v_w$ is given by:

\[v_w = \frac{f_e a_v}{f a_s} \]

From this, we can calculate the width of the view in the scene, $w$, at a given depth, $d$:

\[w = v_w \frac{f}{d}\]

Finally, the conversion between pixels to distance is:

\[dist = pix \frac{w}{r_h}\]

\begin{figure}[h!]
        \begin{subfigure}{\columnwidth}
            \centerline{\includegraphics[width=\textwidth]{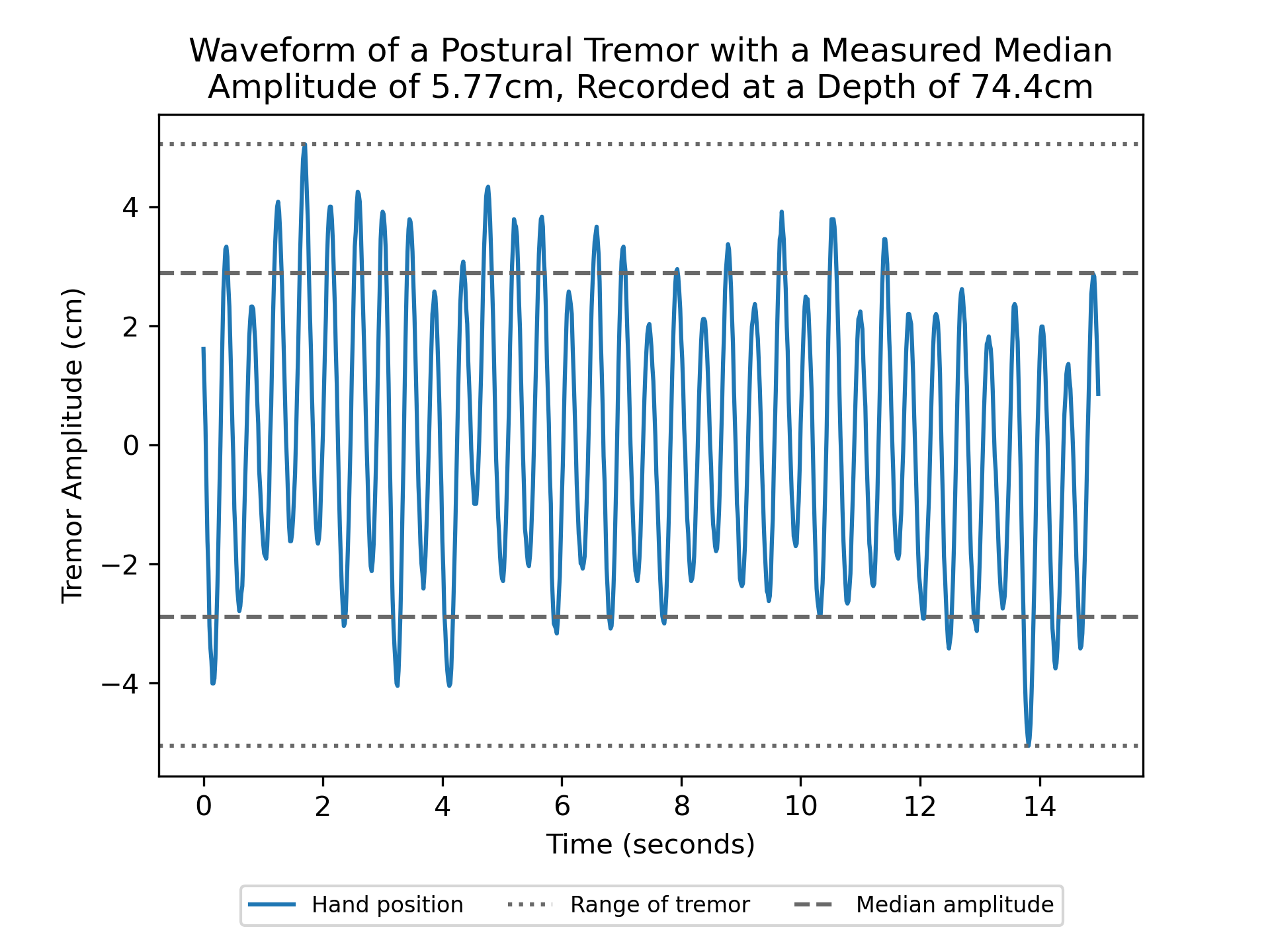}}
            \caption{}
        \end{subfigure}
        \begin{subfigure}{\columnwidth}
            \centerline{\includegraphics[width=\textwidth]{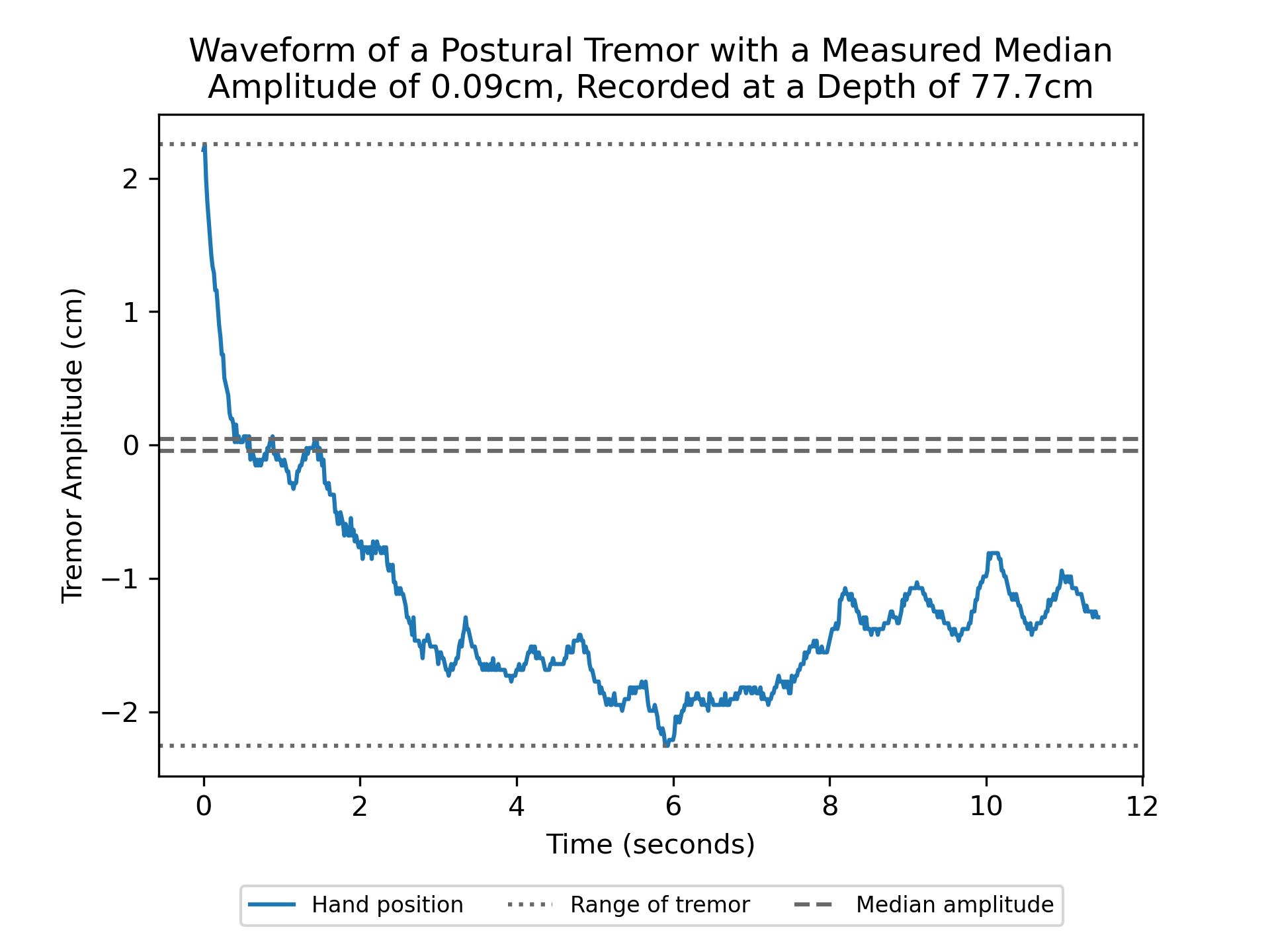}}
            \caption{}
        \end{subfigure}
        \caption{ (a) shows a 'typical' tremor signal with a consistent dominant frequency. The median peak-trough distance is 5.77 cm. (b) shows gross translational movement over time that does not correspond to tremor, which is correctly ignored when amplitude is calculated.}
        \label{fig:tremor-signal}
\end{figure}

Code for extracting the depth measurement and for computing the amplitude is available at \url{https://github.com/jamesbungay/cv-tremor-amplitude}. The output of the entire process, for two example waveforms, is shown in Figure \ref{fig:tremor-signal}. In figure \ref{fig:tremor-signal}a, the dominant frequency of the tremor is visible, and the median peak-trough distance is 5.77 cm. In contrast, Figure \ref{fig:tremor-signal}b shows an example in which there is a gross change in $x$ over the 12 second video recording, but there is no high frequency oscillation caused by tremor. In this case, the method correctly determines that there is no meaningful tremor (Median tremor amplitude = 0.09 cm)

\subsection{Method Validation}
We undertook a methods agreement analysis to assess the performance of the tremor amplitude algorithm. No participants were recruited; all data were fully anonymous self-recordings. Given this, the University of Manchester advised that local ethics was not required.

\subsubsection*{Data Collection}
\begin{figure}[htbp]
    \centerline{\includegraphics[width=0.4\textwidth]{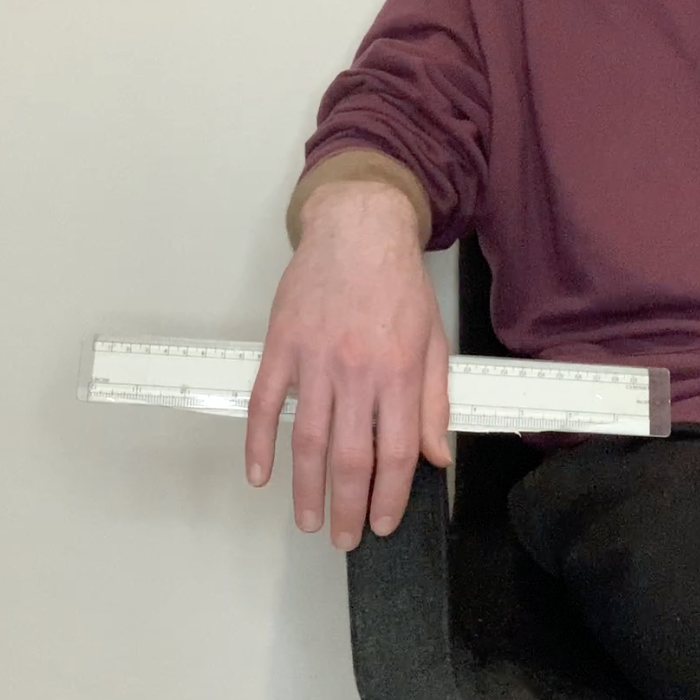}}
    \caption{A research participant in position to measure resting tremor.}
    \label{fig:phase-3-rest-tremor}
\end{figure}

We recorded a set of videos of two members of the study team (JB, OE) Videos were recorded using an iPhone XR smartphone at 1080p resolution and at 60 fps. The smartphone was attached to a tripod, and we ensured that the videoed area was lit well using a ring light. A ruler was placed directly behind the hand. This allowed us retrospectively to measure the tremor amplitude from the video recording.

We simulated two common types of tremor. Resting tremor was simulated by the subject resting their forearm on a chair arm and rotating their wrist to create side-to-side motion as shown in Figure \ref{fig:phase-3-rest-tremor}. Postural tremor was simulated by the subject raising an outstretched arm parallel to the floor, and with the thumb closest to the camera. The subject made oscillatory hand movements up and down. The camera was oriented so that the principal direction of tremor was horizontal, with respect to a portrait video frame.

For both resting and postural tremor, we simulated tremor amplitudes according to five categories (No tremor, small ($<$1cm), medium ($\approx$2cm), large ($\approx$5cm) and very large ($>$10cm), which correspond to the tremor categories used within both the Unified Parkinson's Disease Rating Scale and Essential Tremor Rating Assessment Scale \cite{goetz2008movement, elble2016essential}. Videos were recorded at three depths {50 cm, 75 cm, 100 cm}. The two team members had skin tones of II and VI on the Fitzpatrick scale. In total $2 \times 2 \times 5 \times 3 = 60$ videos were recorded.

We recorded an additional set of 8 videos to assess how well hand detection worked under various types of occlusion. For both subjects, we recorded videos of each wearing a ring, a plaster on the dorsum of the hand that simulated having an accelerometer strapped to the hand, and a hand with fewer than five fingers showing.

\subsubsection*{Data Analysis}

We compared the gold-standard ruler-measured amplitude with our computer vision approach using Bland-Altman agreement analysis \cite{bland1986statistical}. The output of this is bias, which is the mean difference between the computer vision and gold standard and 95\% limits of agreement (LoA), which may be regarded as the maximum difference between the two methods for 95\% of future measurements. In subgroup analysis, we assessed whether there were differences in the bias and LoA for skin tones II and VI using a t-test. 

The additional videos with occlusion were analysed qualitatively in two stages. First, we assessed whether the hand tracker could correctly and reliably identify the salient points. Second, we compared the calculated tremor amplitude against the ruler-based amplitude measurement. Bland-Altman plots were not plotted in these cases, as the small number of videos would mean that the plots would be meaningless.

\section{Results}

\begin{figure}[htbp]
    \centering
    \includegraphics[width=0.5\textwidth]{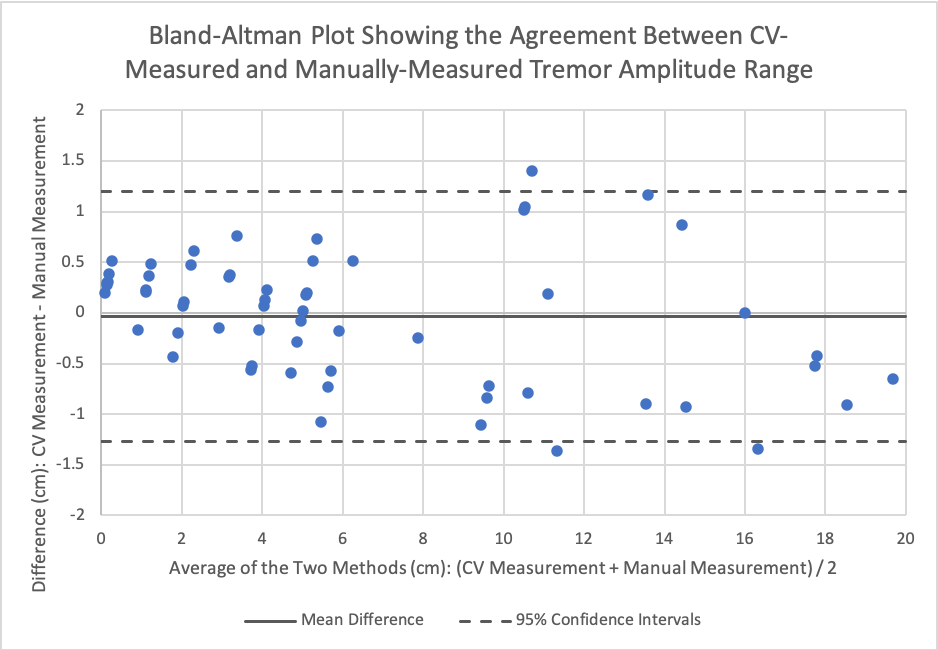}
    \caption{Bland-Altman plot of CV-measured and manually-measured amplitude range.}
    \label{fig:results-bland-altman}
\end{figure}

Bland-Altman analysis showed a mean difference of -0.04cm, with 95\% LoA of -1.27cm to 1.20cm; the associated Bland-Altman plot is shown in Figure \ref{fig:results-bland-altman}. There was no meaningful or statistically significant difference in bias and LoA when the cohort was split according to Fitzpatrick score.

The high limits of agreement can be explained by the manual measurements having a low precision of $\pm1$ cm. This low precision resulted from two factors. First, the resolution of some of the videos was insufficient to be able to discern the millimetre markings on the ruler. Second, parts of the ruler were occasionally obscured from view by the hand. Measurements thus had to be taken by extrapolating neighbouring ruler markings. Mean difference is not affected by the low precision of manual measurements.

\section*{Discussion}
This pilot work shows, for the first time, a smartphone video method for estimating amplitude of hand tremor. In our controlled tests on simulated tremor from healthy individuals, our method showed minimal bias and 95\% limits of agreement -1.27cm to 1.20cm over a wide range of tremor amplitudes, for two different skin tones. In addition, \textit{ad hoc} tests qualitatively showed this approach to be robust under a range of simulated real world conditions.

Previous research for measuring hand tremor amplitude has used bespoke sensors. The acceleration signal amplitude recorded by wrist-worn accelerometers has been used as a proxy for the true tremor size \cite{aljihmani2019spectral}. In principle, the acceleration signal can be integrated to provide true distance, but the resulting signal is likely to be noisy. Electromagnetic position sensors have been used, and claim a fidelity of 0.45 mm \cite{perera2019validation}. Our approach differs by using common sensor modalities that are readily available on most modern smartphones.

While our results are slightly poorer than some existing methods, we believe that these are mainly due to limitations with our experimental setup. Our method contains three potential sources of error. First, from rounding error due to discretization of distance in pixels. For a modern smartphone camera with high resolution, we can assume this to be negligible. Second, from errors in depth measurement. In local tests, we showed an average depth error of 0.38 cm at true distance 100 cm. Using trigonometry, we calculate that this would correspond to a possible error of up to  $0.38/100 = 0.38\%$. These sources of error are limitations of the camera technology and their sum can be considered a maximum lower bound on error. In addition, a third error, caused by rounding rounding the visual gold standard to the nearest centimeter, leads to an error of $\pm0.5$ cm.

While this pilot work provides proof of principle, it is limited in a few key respects. First, the tremor amplitude is only calculated in the plane of the camera image. While this is sufficient for some clinical scenarios, we know that tremor can be very heterogeneous, depending on clinical condition. For instance, tremor associated with Parkinson's disease is commonly described as a `pill-rolling' tremor, which is characterised by rotation of the wrist. Second, the ruler used to provide a reference amplitude measurement was often occluded by the hand. This meant that we were unable to make accurate reference measurements, instead rounding to the nearest centimeter. This in turn led to unreliable estimates of the true agreement with the video-based method. Third, data collection was undertaken using an iPhone only. We note that most modern smartphones contain at least one method for measuring depth data, and that our approach should therefore be generalisable to other devices. An alternative, where there is no direct method to measure depth, would be to estimate depth via depth-from-motion methods \cite{valentin2018depth}.

To address these issues, we are currently conducting a larger validation study using data from patients with multiple types and acuity of tremor. In this study, we will also investigate whether full 3D depth map videos can improve estimation of tremor amplitude.

\section*{Conclusion}
We have demonstrated a smartphone video approach for measuring tremor amplitude. In conjunction with a method for measuring frequency (see \cite{williams2021accuracy}), this method can be objectively and contactlessly measure the key clinical components of tremor in near real-time The method has potential uses for diagnosis, and remote monitoring of disease progression or drug response. 

\bibliography{main.bib}{}
\bibliographystyle{IEEEtran}

\addtolength{\textheight}{-12cm}   



\end{document}